\documentclass{article}

\usepackage[nonatbib,final]{ap_2020}

\usepackage[utf8]{inputenc} 
\usepackage[T1]{fontenc}    
\usepackage{hyperref}       
\usepackage{url}            
\usepackage{booktabs}       
\usepackage{amsfonts}       
\usepackage{nicefrac}       
\usepackage{microtype}      
\usepackage{biblatex}
\usepackage{multirow}
\usepackage{graphicx}
\usepackage{xcolor}

\title{Alquist 3.0: Alexa Prize Bot Using Conversational Knowledge Graph}

\author{
  Jan Pichl, Petr Marek, Jakub Konr\'ad,
  Petr Lorenc, Van Duy Ta\\
  Faculty of Electrical Engineering, CTU Prague\\
  Prague, Czech Republic \\
  \texttt{\{pichljan, marekp17, konrajak, lorenpe2, tavanduy\}@fel.cvut.cz} \\
   \And
   Jan \v{S}ediv\'{y} \\
   CIIRC, CTU Prague \\
   Prague, Czech Republic \\
   \texttt{jan.sedivy@cvut.cz} \\
}

\begin{document}

\maketitle

\begin{abstract}
  The third version of the open-domain dialogue system Alquist developed within the Alexa Prize 2020 competition is designed to conduct coherent and engaging conversations on popular topics. The main novel contribution is the introduction of a system leveraging an innovative approach based on a conversational knowledge graph and adjacency pairs. The conversational knowledge graph allows the system to utilize knowledge expressed during the dialogue in consequent turns and across conversations. Dialogue adjacency pairs divide the conversation into small conversational structures, which can be combined and allow the system to react to a wide range of user inputs flexibly.
  We discuss and describe Alquist's pipeline, data acquisition and processing, dialogue manager, NLG, knowledge aggregation, and a hierarchy of adjacency pairs. We present the experimental results of the individual parts of the system.
\end{abstract}

\section{Introduction}\label{introduction}
 This paper describes the third version of the socialbot ``Alquist'', a conversational system designed to converse coherently and engagingly with humans on popular topics.
 
 To provide users with an informative and fluent experience, we enhance a novel approach based on creating the dialogue using a conversational knowledge graph. This allows Alquist to work with smaller template-based sub-dialogues, and be more reactive to the broad range of user responses present in natural dialogue. The conversational knowledge graph also allows Alquist to store the knowledge expressed during the conversation and utilize it in following interactions with the user. The system is more modular and provides a unique experience in each session. It can have a seamless conversation on popular topics like pop culture, sport, or technology.  We also made a significant increase in the performance of our NLU module using new state-of-the-art technologies, and focused on joining more models into one to maintain accuracy and decrease memory requirements. Additionally, Alquist 3.0 can be enhanced with the approach presented in the paper about the second version of Alquist \cite{alquist2}. Such a combination offers more vibrant conversations to the users.

The Alquist socialbot builds on the experience and knowledge gained from Amazon Alexa Prize '17 and '18. Previous results showed that users could have a meaningful conversation with a socialbot, but it needs to be more reactive and accurate in the answer. These facts motivated the changes which we introduced into the third version of the socialbot Alquist.

The structure of this paper is divided into several sections. First of all, we focus on our novel approach based on the knowledge graph. After this, we dedicate a section to the overall system architecture, highlighting essential parts in subsections. Finally, we provide example conversations, we describe the experiments we performed during the creation of the system, and we summarize our findings.

\section{Motivation}\label{motivation}
Let us start by explaining why we decided to use the architecture described later in the paper. The architecture of the system has always been designed to address two contradictory requirements. The first requirement is to be in control of the dialogue flow during the development process. The second one is to create such dialogues that can respond to various and unexpected user inputs.

The main part of the first version of our system \cite{alquist1} developed during the Alexa Prize 2017 was created using a complex dialogue structure, which we called ``dialogue automata''. These structures were supposed to deliver engaging conversations thanks to a rich dialogue structure, which can present all the relevant information to the users as they traverse through it. While this approach works very well for cooperative users as the structure guides them through the conversation, it can be easily broken by the users asking additional questions or trying to change the direction of the dialogue. Additionally, many of the dialogue nodes could remain unvisited as the unexpected input messages did not let the bot finish the dialogue.

To solve the problems mentioned above, we introduced the second version of the system \cite{alquist2} during the Alexa Prize 2018. We introduced smaller structures called sub-dialogues, which can be chained dynamically, allowing the users to have different dialogue experiences. Each of the sub-dialogues has its Dialogue Management model, and the high-level decision of which one should be triggered is made by a combination of the results of the Topic Switch module and the Intent module. This new approach reduced the problem, yet it was far from a complete solution.

We run analyses of the errors produced by the system. We manually went through 100 randomly selected conversations and marked the turns which we found incoherent (the engagement of the system was not evaluated). Note that the evaluation was done on a subjective basis. For each error turn, we marked which of the system components caused the erroneous response (some turns can contain multiple errors). The distribution of the errors is shown in Table \ref{errors}. An error is marked only for those turns that produced a response violating the coherence of the dialogue. There can be, for example, a turn with a punctuation error but the responses are still generated correctly---such a turn is not marked as an error.

\begin{table}[h]
\caption{Error distribution in the second version of Alquist. The analysis was done manually on 100 conversations. Some of the analyzed turns contained multiple errors, the most typical combination was an error in the Topic Switch module and in the Intent module. Conversation examples are synthesized by our team based on observed patterns in the data.}
\label{errors}
\begin{center}
   \begin{tabular}{c|c|l}
    \hline
    & Occurrence & Example \\ \hline \hline
    \multirow{3}{*}{Topic Switch} & \multirow{3}{*}{44 \%} & \textbf{Bot: } \ldots How is it going?\\
     &  &  \textbf{User: } it's going good what happened today in history\\
     &  &  \textbf{Bot: } \textcolor{red}{Well, I'm always busy as people keep chatting with me.} \\\hline
    Intent & 28 \% & \textbf{User:} how old are you (intent: \textcolor{red}{how\_are\_you})  \\ \hline
    Punctuation & 19 \% & \textbf{User:} good I'm \textcolor{red}{<SEP>} good you're really good  \\ \hline
    Dialogue Act & 6 \% &  \textbf{User:} who's boyfriend (DA: \textcolor{red}{Other}, correct Wh-question) \\ \hline
    \multirow{3}{*}{Dialogue Manager} & \multirow{3}{*}{4 \%} & \textbf{Bot: } \ldots Who do I have the pleasure of speaking with?\\
     &  &  \textbf{User: } why are fire trucks red\\
     &  &  \textbf{Bot: } \textcolor{red}{Hey Red, I'm happy to meet you!}\\\hline
    \end{tabular} 
\end{center}

\end{table}

In pursuit of mitigating the errors above, we came up with the third version of our system, which we describe in this paper. We present a concept of adjacency pairs that can be chained flexibly. Thanks to the fact that dialogues built using adjacency pairs are short, we do not need the Topic Switch component, which caused most of the errors. Moreover, we use a conversational knowledge graph that stores the information available for use as well as the information expressed by the users in previous sessions. The previous versions of Alquist were also able to store users' preferences; however, only in predetermined parts of dialogues. With the new approach, the system can remember facts expressed at any point. The utilization of the knowledge graph allows us to remember a significantly large number of facts extracted during the dialogue, which can be subsequently used in the interaction by Alquist, or asked about by users.

\section{System architecture}
The Alquist 3.0 system architecture is conceptually similar to the architecture of previous versions of the system. It is based on the Knowledge Graph and the main component Pipeline, to which additional micro-services are connected.

\subsection{Conversational Knowledge Graph}\label{knowledge_base}
To be able to handle more complex interactions with the user, it is necessary to have access to real-world knowledge and objective factual information as well as personalized (subjective) knowledge, such as opinions, preferences, and others. To represent the knowledge and opinions of our system about the world, we have created an RDF graph database where we used a dump of the Wikidata database\footnote{https://www.wikidata.org/wiki/Wikidata:Database\_download} as its base.

We have also designed a custom ontology that is to be partially mapped to the ontology used by Wikidata. Our ontology is inspired by other common ontologies such as Schema.org\footnote{https://schema.org/} and EVI. However, we expanded it with objects representing the user and the bot. We also added properties modeling personalized relationships between the user or bot and various other objects in the database. Examples of such properties are \emph{likes, hates, hasFavorite, hasOpinion}.

\begin{figure*}[ht]
\begin{center}
\includegraphics[width=0.8\linewidth]{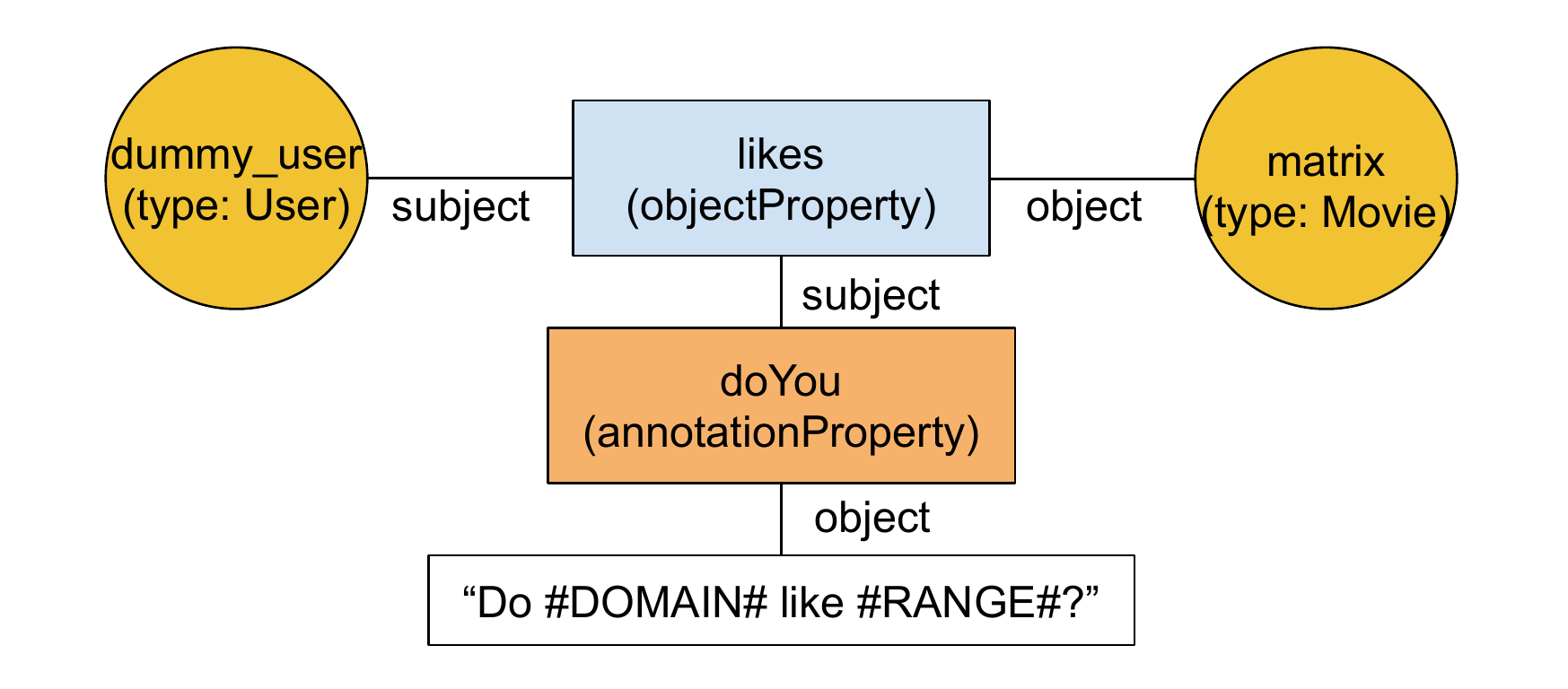}
\caption{Illustration of sentence structure examples as annotation properties} 
\label{fig:kb}
\end{center}
\end{figure*}

As shown in Figure \ref{fig:kb}, we also created custom annotation properties in the ontology (e.g. \emph{doYou}). These properties are tied to object properties. Their range is commonly a string (e.g. \emph{``Do \#DOMAIN\# like \#RANGE\#''}), and they represent delexicalized examples of sentence structures that represent the domain object property. This allows us to do two things. We can automatically generate and update datasets for dialogue acts detection directly from the knowledge base (KB). Furthermore, we can use the sentence structures assigned to properties in adjacency pairs templates to generate system responses regarding relationships between objects in the database. 

Additional examples of sentence structures, such as:
        \begin{itemize}
            \item \emph{ProvideInformation\_Negative} - \#DOM\#'s name is not \#RAN\#.
            \item \emph{OpenQuestion\_Object\_Positive} - What is \#DOM\#'s name?
             \item \emph{YesNoQuestion\_Positive} - Is \#DOM\#'s name \#RAN\#?
        \end{itemize}
        
The new approach to representing knowledge allows the system to create a profile of the user directly in the knowledge database, based on the information the user shares with the system. The system is then able to reference the profile during the conversation, and this leads to more variability and higher personalization. However, this personalized engagement needs to work in both directions – and so the profile of the bot is represented in the knowledge base in the same way. The database contains a representation of the bot, complete with its preferences, personal profile, likes, and dislikes. The profile is adjusted and anonymized for the purposes of the competition, but the system can draw information from it in real-time and include it in the conversation. 

\subsection{Pipeline}
The fundamental part of the system responsible for message processing is the Pipeline, which handles communication between the Natural Language Understanding (NLU), Knowledge Graph, Dialogue Management, and Natural Language Generation (NLG) components during the process of response creation.

The input to the pipeline is the message with the highest confidence score produced by the Alexa ASR. The message is sent to the NLU component, which segments the message. A segment is a part of the message representing one semantic unit. The NLU component also provides entity and property annotations for each segment (see section \ref{NLU}). Alquist generates a set of actions for every segment, based on the NLU annotations in the next step (see section \ref{action}). Dialogue management creates a response for each action based on the adjacency pair, which is assigned to the action (see section \ref{dialogueManagement}). The NLG component adds entities extracted from the knowledge graph into the responses and converts them into syntactically correct forms (see section \ref{lexicalize}).
After this step, Alquist selects one action for each segment, which will produce a response based on NLU confidence scores and additional criteria (see section \ref{actionSelector}). Additionally, to have a more engaging and coherent conversation with the user, the system also enhances the response by a fun fact and a question based on the annotations of the response (see section \ref{fun-fact}). After this step, the final response is presented to the user.

\begin{figure*}[ht]
\begin{center}
\includegraphics[width=\linewidth]{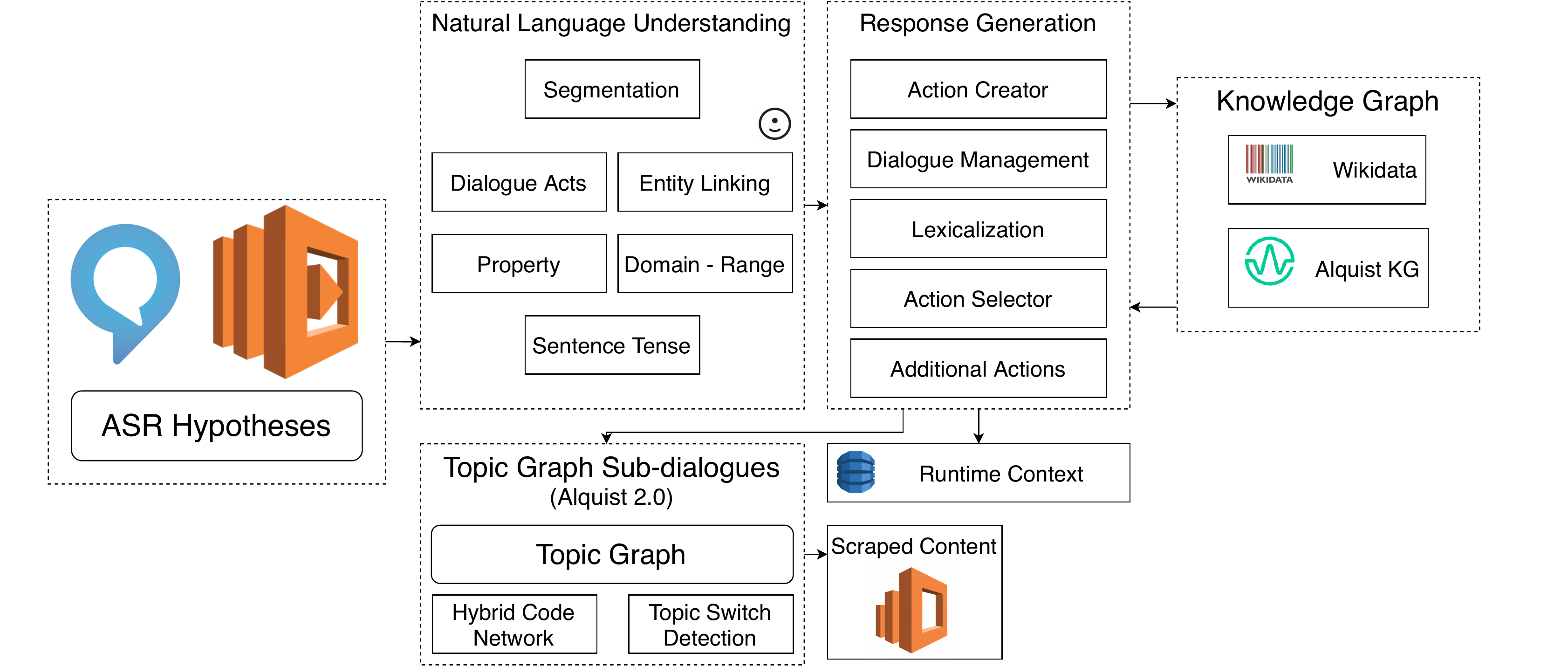}
\caption{The system architecture schema} 
\label{fig:architecture}
\end{center}
\end{figure*}

We summarize the pipeline in Figure \ref{fig:architecture}.

\subsection{Natural Language Understanding}\label{NLU}

The Natural Language Understanding component is a set of analysis machine learning models producing the annotations essential for further processing in the pipeline. These models can be triggered in an arbitrary order. An exception is the sentence segmentation, which needs to be triggered first, as the remaining models annotated each segment separately. The user message is processed as follows. The input for the NLU consists of $ n $ speech recognition hypotheses, each hypothesis is processed with the segmentation model first, and each segment from a hypothesis is used as the input for the additional models. 

\subsubsection{Sentence Segmentation}

The goal of this model is to split the sentences into clauses. The standard way how to tackle this task is to decide based on punctuation marks. However, our input consists of speech hypotheses without punctuation. Therefore, we start by using punctuation restoration, and afterward, we split the sentences based on the restored punctuation marks.

Compared to the former Alquist system where we used a bidirectional neural network (TBRNN \cite{tilk2016bidirectional}), we decided to use a BERT based model \cite{devlin2018bert}. Our model consists of a pre-trained BertForMaskedLM layer and a linear layer that predicts the punctuation mark after the middle word of a fixed size input sequence.

Since the model predicts only the punctuation after the word in the middle, we need to add padding to the beginning and the end of the input speech hypothesis. Afterward, we feed the model with sequences of a fixed segment size based on the moving window, and the model returns one of the possible punctuation characters for each word in the original speech hypothesis.
 
The results of the new model are shown in Table \ref{table:segmentation_experiment}. We can see a significant improvement in all metrics compared to the older model. Moreover, since the BERT model is more computationally demanding, we also analyzed the performance of various BERT distillations \cite{sanh2019distilbert, lan2019albert}. Based on the observed results, we can conclude that all tested models outperform the former bidirectional one. 

\begin{table}[!htb]

\begin{center}
\caption{Results of the experiments on IWSLT 2012 English dataset.}
\scalebox{0.9}{%
\begin{tabular}{l|ccc|ccc|ccc|ccc}
\hline
\textit{Model} & \multicolumn{3}{c|}{\textit{Comma}}                    & \multicolumn{3}{c|}{\textit{Period}}         & \multicolumn{3}{c|}{\textit{Question}}        & \multicolumn{3}{c}{\textit{Overall}}         \\
           & Pr.  & Re.  & F\_1 & Pr.  & Re.           & F\_1 & Pr.           & Re.  & F\_1 & Pr.           & Re.  & F\_1 \\ \hline \hline
TBRNN \cite{tilk2016bidirectional}     & 65.5 & 47.1 & 54.8 & 73.3 & 72.5          & 72.9 & 70.7          & 63.0 & 66.7 & 70.0          & 59.0 & 64.4 \\
BERT \cite{devlin2018bert}  & \textbf{74.9} & \textbf{66.8} & \textbf{70.7} & \textbf{81.8} & 85.5 & \textbf{83.6} & 66.7 & \textbf{73.9} & \textbf{70.1} & 74.4 & \textbf{75.4} & \textbf{74.8} \\
DISTILBERT \cite{sanh2019distilbert} & 70.3 & 58.6 & 63.9 & 81.4 & 76.3          & 78.8 & \textbf{78.4} & 63.0 & 69.9 & \textbf{76.7} & 66.0 & 70.9 \\
ALBERT \cite{lan2019albert}    & 84.0 & 44.1 & 57.8 & 77.0 & \textbf{87.6} & 82.0 & 64.6          & 67.4 & 66.0 & 75.2          & 66.4 & 68.6 \\ \hline

\end{tabular}%
}
\label{table:segmentation_experiment}
\end{center}
\end{table}

\subsubsection{Dialogue Act detection}

Since one of the key concepts in the pipeline is the processing of adjacency pairs (as described later in this paper), we need to identify the dialogue act in the message of a user. We created a hierarchical structure of the dialogue acts. We treat this problem as a simple text classification task where each class is defined by the combination of the tags on the path from the tree root to a leaf node. You can find a description of the classes in Appendix \ref{ap:da}. We defined in total 37 leaf classes, each of the class can be mapped to one or more dialogue act class presented in the Switchboard dataset\footnote{\url{https://web.stanford.edu/~jurafsky/ws97/manual.august1.html}} \cite{godfrey1992switchboard}.

The data for the dialogue act detection model were created manually through a template mechanism which we have been using since the previous version of our system. The created data are stored in the KG. We adopt the model architecture which we have been using for intent detection and dialogue act detection in previous version of Alquist \cite{alquist2}, which is inspired by \cite{kim2014convolutional}.

\subsubsection{Entity Linking}

As our new approach is based on working with a knowledge graph (KG), we need to identify each entity with its unique ID to be able to generate a proper response. This task is commonly referred to as entity linking.

Entity linking, as stated in \cite{broscheit2020investigating}, consists of three parts: mention detection (MD), candidate generation (CG) and entity disambiguation (ED). We tackle each problem as a separate unit to be able to leverage Cobot KG.

To generate mentions (text spans of potential entity occurrence), we used neural network based on bidirectional LSTM model, where the training set was manually labeled from a small subset of the user's messages. We test our model on the sequence labeling task, see the result in Table \ref{tab:entity_linking_sequence_labeling}.

After retrieving the possible text spans, we combine the candidate generation  with entity disambiguation. First, we have to decide if the entity is from our ``private'' knowledge graph or from the ``general'' knowledge graph (for example, Wikidata). We trained our neural network model to classify into two label types. If the model returns the label classifying the entity candidate as one from our ``private'' KG, then we have a set of manually designed rules to decide which possible entity should be assigned to that specific text span. The other case is that the candidate is from the ``general'' KG: then we query an Evi knowledge graph service to make the mention resolution. We collect all candidates for possible text span that the system returns, and query Evi knowledge graph service again to get their IDs (more specifically, Wikidata ID).

Further processing is coupled with property detection (see section \ref{property}), where we map the text span to two classes – domain or range. We have to match predictions from property detection and entity linking together to be able to decide which entity (linked to the knowledge graph) is the domain and which one is the range. Our further processing of user queries is principally dependent on these annotations (see \ref{action}).

{
\begin{table}[h]
\centering
\caption{Results for sequence labeling}
    {\renewcommand{\arraystretch}{1.2}
    \tabcolsep=.3cm
\begin{tabular}{c|c|c|c|c}
\hline
Model & Dataset & Precision & Recall  & F1 \\
\hline \hline
Bi-LSTM & CoNLL 2003 & 90.0 & 89.4  & 89.8 \\ \hline
Bi-LSTM & Own manually labeled data & 93.6 & 92.0 & 92.6 \\
\hline

\end{tabular}
}
\label{tab:entity_linking_sequence_labeling}

\end{table}
}

\subsubsection{Property detection}\label{property}

The knowledge graphs stores facts as triples, with each triple containing two entities and one relation. The user utterance can be mapped to this triplet – the entities are called domain and range; the relation between them is called property. To properly handle the response, we need to retrieve the property out of the user's message.

We tackle the problem of property detection as a multi-task model. These models are showing promising results on a task like joint slot-filling and intent classification \cite{att-based}.

Based on \cite{bimodelRNN}, we design our neural network with the novel approach based on the multi-task loss function. The model is trained on two tasks – sequence labeling and sentence classification.

 To find the most suitable word embeddings on our proposed system, we used three types of word representation as a vector of floats:
 
   \begin{itemize}
      \item GloVe \cite{glove} - 300d
      \item Fasttext \cite{DBLP:journals/corr/BojanowskiGJM16} - 300d
      \item Bert \cite{devlin2018bert} - 768d - without fine-tuning
  \end{itemize}
  
We measure our metrics on the publicly available ATIS dataset provided by Microsoft\footnote{Available at github.com/Microsoft\newline/CNTK/tree/master/Examples/LanguageUnderstanding/ATIS}. The data contain spoken utterances classified into one of 26 intents. Each token in a query utterance is aligned with IOB labels. Primarily, the dataset is used for intent recognition and slot filling, but it is sufficient for our testing purpose. The results are shown in the Table \ref{tab:atis-model1-ner-embeddings}.

\begin{table}[h]
    \begin{center}
    \caption{Results on ATIS dataset}
    {\renewcommand{\arraystretch}{1.2}
    \tabcolsep=.8cm
        \begin{tabular}{c|c|c}
        \hline 
        \textbf{Embeddings} & \textbf{Intent F1 accuracy} & \textbf{Slot-filling F1 accuracy}  \\\hline \hline
        BERT & 92.45 & 95.78  \\\hline
        fasttext & 90.70 & 96.83 \\\hline
        GloVe & \textbf{95.74} & \textbf{98.44} \\\hline
        \end{tabular}
    }
    \label{tab:atis-model1-ner-embeddings}
    \label{tab}
    \end{center}
\end{table}
    
From the result, we can see that GloVe embeddings outperform the others. Unfortunately, they have the disadvantage of the limited possibility of dealing with out-of-vocabulary words, so their usage is limited. We will do more research about fine-tuning the BERT, which also shows promising results.

To create a more robust model, we used the WebNLG dataset \cite{webnlg} as an additional source of data for training. It consists of a set of triples extracted from DBpedia, where the text is the verbalization of these triples. We mapped the properties to follow the Wikidata taxonomy.

The final model of property detection has two outputs. The sequence labeling output will classify each token into three categories – domain, range, or outside. The text classification output predicts the label of property (mapped to our KG) which is mentioned in the user's message.

\subsubsection{Grammatical tense classification}

Because the user is speaking naturally and we need to understand what is the time-span of the event which he might be talking about, therefore we classify an utterance into the correct tense. This sub-component uses the morphological structure of the word and part-of-speech tagging designed in spaCy \cite{spacy2}. SpaCy's implementation is based on rules and can predict the present and past tense. For the future tense, we develop a set of own rules.

\subsection{Action Creator}\label{action}
Alquist has to react to the user's message appropriately. The input can be noisy either due to the ASR or because of the user over- or under-specifies the facts in the message (we observed over-specification of facts being a bigger problem for socialbots). For these reasons, sentence segmentation divides the input message into several segments, and each segment is processed by NLU independently. The NLU component outputs several classification hypotheses for each segment. This leaves the system with a large amount of NLU classification hypotheses, from which it has to choose the relevant ones for the creation of the response. We define \emph{action} as a possible way, how to react on the user query, so the Action creator is the first component of the pipeline contributing to the process of response creation. Its task is to select all combinations of NLU classification hypotheses (entity types, properties, and dialogue acts), which are proposed for further processing.

There is only a finite number of combinations of NLU classifications meaningful for processing. The meaningful combinations are closely tied to the structure of the knowledge graph. We create actions only for those combinations of NLU classifications that contain an entity of a certain type, a certain property, and a certain dialogue act. For example, we create an action containing the entity type $ chatbot $, the property $ name $ and the dialogue act $ open $ $ question $. This combination represents the question ``What is your name?''

The Action Creator creates several actions for each sentence segment, yet it does not create actions that do not respond to the structure of the Knowledge Graph. This fact allows Alquist to take advantage of more than just the Top-1 NLU classification. Thus, it processes potentially noisy messages in a computationally feasible way.

\subsection{Dialogue Management}\label{dialogueManagement}
\begin{figure*}[ht]
\begin{center}
\includegraphics[width=0.5\linewidth]{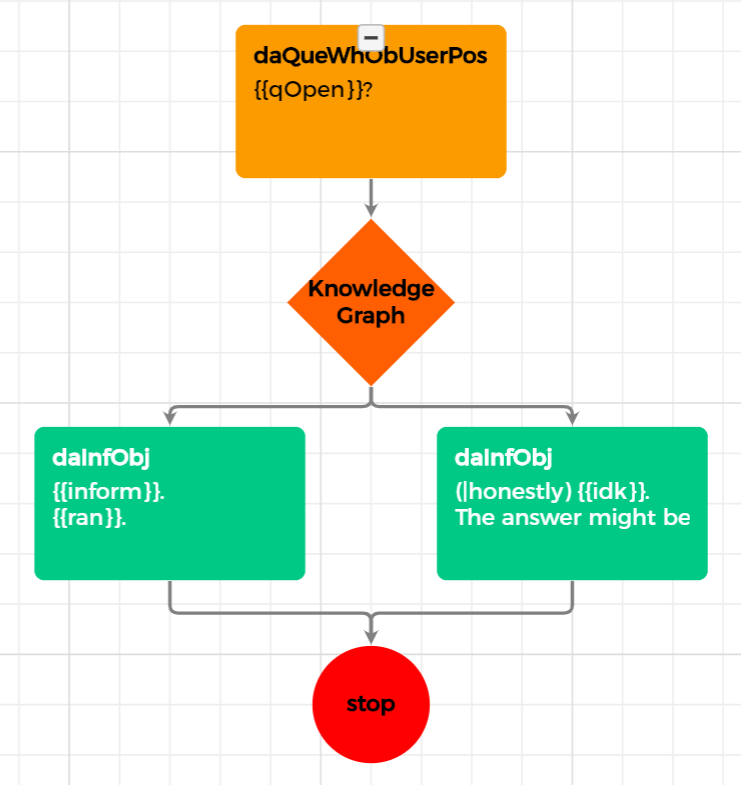}
\caption{Adjacency pair representing an open question. The pair is started by the user with the $ open $ $ question $ dialogue act. The adjacency pair queries the knowledge graph first. Based on the result of the query, it informs about the result (in the left branch), or it apologizes for not knowing the answer (in the right branch).} 
\label{fig:adjacencyPair}
\end{center}
\end{figure*}

Alquist has to decide what response it will produce for each action. This decision is made by the Dialogue Manager. The Dialogue Manager creates a response based on actions created by the Action Creator, and existing adjacency pairs. Adjacency pairs \cite{schegloff1973opening} are a linguistic concept according to which adjacent utterances in dialogue often form predetermined pairs. We utilize the concept of adjacency pairs in the form of simple dialogue structures, which are small and generic. Thanks to their size, adjacency pairs can be chained together and thus can flexibly react to unexpected input messages. Because of their generality, they can cover large parts of the knowledge graph with a small number of universal dialogue structures. An example of adjacency pairs handling open questions is shown in Figure \ref{fig:adjacencyPair}.

It would be very time-consuming to create specific responses for all adjacency pairs. Often, the same adjacency pair will deal with very similar responses concerning different properties – in such situations, the responses usually differ in a single word. For this reason, we do not create the specific adjacency pairs directly. Instead, we introduce three levels which are utilized during the creation: a word level, a sentence structure level, and an adjacency pair level. 

 The first level operates with words. We selected a set of word categories, which describe the properties. The categories, for example,  are Verb, Subject, or Pronoun. It is the task of the person working on certain properties to select the appropriate word for each category.
 
 The selected words are used as input for the level of sentence structures. Each structure is represented by a pattern of words. These patterns constitute sentences that express a certain meaning. The sentence structures can, for example, inform about the value of the property or ask about the value of a property. 
 
 The sentence structures are finally utilized in the adjacency pairs. An adjacency pair in our implementation consists of nodes and edges. The nodes represent the responses, and edges represent transitions between them. The response contains types of sentence structures, which are substituted by their real values depending on the property for which the adjacency pair is being utilized.
 
 As shown, the word, sentence, and adjacency pair levels decrease the amount of work necessary for the design of responses. 

\subsection{Lexicalization}\label{lexicalize}
The knowledge graph contains a large amount of knowledge which relates to many different entities, individuals, or events happening at different points in time. If we want to describe such knowledge in natural language, we have to express all of these variables with the correct word morphology and sentence syntax. 

However, it is intractable to create responses using all possible word forms and sentence syntaxes to encapsulate all entities or events contained in the knowledge graph. For this reason, Alquist uses the Lexicalization module, responsible for converting responses into the correct form.

The Lexicalization module has four tasks. It fills the values of entities in place of delexicalized tokens, it switches pronouns based on the gender of entities, it converts nouns into singular or plural form, and it converts the verb into the appropriate tense.

The response created by the dialogue manager contains delexicalized tokens $ Domain $ and $ Range $. We can see the example in the response \textit{``DOMAIN loves to talk about RANGE''}. Actual entity values extracted from the knowledge graph replace $ Domain $ and $ Range $ tokens, creating the response \textit{``KAREL ČAPEK loves to talk about ROBOTS''} for example. It often happens that the substituting entities are User or Alquist. In such cases, Lexicalization applies an additional rule: if the $ Domain $ or $ Range $ is equal to User or Alquist, the system replaces the delexicalized entities by pronouns $ I $ or $ you $. The part of Lexicalitzation responsible for entity substitution allows Alquist to talk about any pair of entities using a single response.

When the same entity is referenced multiple times, it is natural to reference it by a pronoun and not its name. However, it is not trivial to select the correct pronoun because it depends on the gender and the grammatical number of the entity. This is the task of the second part of lexicalization.
The first step determines the grammatical number of an entity based on the value stored in the knowledge graph. If the number is plural, Alquist uses pronoun $ they $. If the number is singular,  Alquist determines the gender of the entity from the knowledge graph. Based on the result, it selects the pronoun $ he $, $ she $, or $ it $. The process of pronoun lexicalization is possible thanks to the utilization of the knowledge graph, which stores the grammatical number and the gender of entities.

The knowledge graph also stores the quantity of entities. An example could be that the user has $ two $ sisters. The delexicalized response for such a fact looks like this: \textit{``DOMAIN have RANGE sister.''} The form of the noun sister depends on the value of a range. If the range is larger than one, the system creates a plural form of the word according to lexical rules. This part of lexicalization allows Alquist to talk about amounts stored in the knowledge graph.

Some of the knowledge stored in the knowledge graph or mentioned during the conversation can have a certain temporal relation. It also influences the response, mainly its verb. The fact that the user \textit{went} to a university is an example of knowledge having a temporal relation. If Alquist wants to continue in a deeper conversation about such a fact, it has to take into account the temporal information. The user used past tense, meaning that he most probably does not study anymore. If this is true, the follow-up question \textit{``What university do you study?''} is not natural. Instead, lexicalization works with the tense used in the user's message and changes the verb tense in the response accordingly, using lexical rules. Thanks to this lexicalization system, Alquist can talk about past, recent, and future events as well.

The main innovation of Alquist 3.0 is its ability to carry flexible conversation grounded by the facts stored in the knowledge graph. This puts a high demand on the syntax of the responses. The syntax needs to encapsulate relations between entities, their amounts, or the time of events. It is labor-wise challenging to create all syntactic variants for responses, considering the large number of all possibilities. Thus it is evident that the lexicalization module, which modifies the syntax dynamically, is essential for a socialbot of Alquist's scale.

\subsection{Action Selector}\label{actionSelector}
Due to the expected noisy input message, the NLU part of the system segments the message into several parts, for which NLU produces several hypotheses of entity recognition and dialogue act detection. In the following step, the Action Creator creates actions out of these hypotheses. This leads to several actions for each segment. Dialogue Manager and Lexicalization discard some of the actions; however, Alquist has to select the most appropriate combination out of the rest. This is the task of the Action Selector.

The main criteria for the selection of actions are combined confidence scores of the NLU classification contained in action. The goal is to select one action for each segment, which maximizes the total score of all selected actions. These criteria select the actions that Alquist is the most confident of, given the noisy input message.

There are also additional constraints to the selection. The selected combination cannot contain actions that lead to the same adjacency pair. The combination also cannot include actions leading to multiple adjacency pairs whose responses consist of questions. The first constraint makes sense, given the desired naturalness of the conversation. It would be unnatural to repeat the same sentence multiple times during a single response. The second constraint improves the coherence of the conversation. We observed in the previous versions of the system \cite{alquist1, alquist2} that asking the user a single question leads to a more prolonged conversation. Thus Action Selector improves the naturalness and coherence of the conversation even when Alquist receives noisy input messages.

\subsection{Generating additional actions}\label{fun-fact}
In previous sections, we described how we generate actions that directly respond to the user's message. We call this action type the \textit{handle action}. However, in order to keep the conversation engaging and exciting for the user, it is necessary to enrich it by injecting additional information, thoughts, and opinions, as well as prompting the user to continue the conversation with additional questions.

We define two additional types of actions that are generated for each segment of the user's message, \textit{fun-fact actions} and \textit{forward actions}. The fun-fact action is the part of the system's response that contains a statement with information related to the entities in the user's message. Forward actions are questions that are again related to the entities in the user's messages, and they ask users for more information or their opinions and feelings about the conversational topic.

Both types of actions are generated using the entities recognized in the segment and the Conversational Knowledge Graph. Properties and entities in the database had a popularity metric assigned to them, starting at 1 with specific properties and entities having their popularity increased manually to make their appearance more common during the conversation. The popularity is then further increased whenever the property or entity is mentioned in the conversation by the user or system. When generating the fun-fact and forward actions, the system generates a large number of possible actions. The popularity of each action is computed as the average of popularities of contained entities and properties. The action with the highest popularity is selected. This system is designed to prevent the bot from introducing conversational topics based on properties and entities that would not provide an exciting conversation.

\section{Latency improvements}

In addition to the coherence and engagement criteria, system latency is another important aspect. It is essential to keep latency below a reasonable level in order to maintain a seamless conversation experience. During development, the NLU component showed to be the most challenging in keeping its latency under the desired level. As we want to keep the overall latency under 500ms (on average), we needed to apply several improvements of the NLU models to comply with this level.

Adding a new model to the NLU module, especially the BERT-based segmentation model, started to degrade the response time of the whole module. We selected constant input to see the performance changes during development. The input was selected to contain multiple segments (\textit{''hello how are you what's your name''}).
The measurements were done on the same instance type as the production system uses.

The NLU version used in Alexa Prize 2018 had the average latency (over 1000 measurements) 153 ms. As we implemented models required for the new architecture, including BERT-based segmentation, the latency grew to 835 ms on average. The first improvement was to change the instance type from the CPU instance to the GPU instance, improving the average latency to 421 ms. However, the GPU instance introduced a latency caused by transferring data from the system memory to the GPU memory – the models process the input immediately, there is no queue to form a bigger batch by combining several requests. Therefore we implemented the following process:
\begin{enumerate}
    \item Run segmentation model on a batch of $ n $ speech hypotheses 
    \item Create input batch from all segment of all hypotheses
    \item Combine all the additional NLU models into one computational graph-–-allowing parallel inference across the models
    \item Run the inference and assign the result to corresponding segments and hypotheses
\end{enumerate}
The combination of batching segments and creating one computational graph from the individual trained models allowed us to decrease the overall NLU latency to 204 ms.

\section{Examples of conversational exchanges}

In this section, we present several dialogue fragments typical for the new aspects of the system. We try to demonstrate multiple parts of the responses as well as the ability to work with knowledge.

\textbf{User:} Do you like music? \\
\textbf{Bot:} Yes, I love music! What music genre is your favorite? \\
\textbf{User:} I really like funk. \\
\textbf{Bot:} I see! \\

\textbf{User:} Alfred Nobel was born in France, right? \\
\textbf{Bot:} No, Alfred Nobel was born in Sweden. Did you know that the synthetic element nobelium is named after him? \\
\textbf{User:} No, I didn't! Thanks for telling me. \\
\textbf{Bot:} No problem. \\

\textbf{User:} What movie is your favorite? And... do you like Tom Hanks? \\
\textbf{Bot:} My favorite movie is Matrix. Yes, I like Tom Hanks. \\

\textbf{User:} Hey, how many siblings do you have? \\
\textbf{Bot:} I don't have any siblings. You have three siblings, right? \\
\textbf{User:} Yes, that's right. \\
(...) \\
\textbf{User:} I have two siblings. \\
\textbf{Bot:} Oh, really? I remembered something else. \\

\textbf{User:} Where was Jara Cimrman born? \\
\textbf{Bot:} I'm sorry but I don't know that... \\
\textbf{User:} Ah, don't worry about it. \\
\textbf{Bot:} Thanks. \\

\section{Conclusion}
Alquist 3.0 brings the innovative usage of conversational knowledge graphs and adjacency pairs. We described how conversational knowledge graphs are essential for open-domain socialbots: it is thanks to the knowledge graph's ability to store the information mentioned during the conversation, as well as to serve as the foundations for constructing dialogues.

We showed that insufficient flexibility of the open-domain dialogue system could lead to lower satisfaction of the users. The utilization of smaller and more flexible dialogue structures can help overcome this issue. We demonstrated how we took advantage of dialogue adjacency pairs, which are small flexible parts of dialogues, how we efficiently create them out of word and sentence structure levels, and how we chain them in order to create flexible, highly coherent, and engaging open-domain dialogues. 

\section{Future Work}

We developed a system that is highly connected to a conversational knowledge graph. The KG contains the information that the system is capable of having a conversation about, as well as relevant information about the user's preferences. The small segments of natural responses allow the system to be more adaptive in the response generation. However, during the development, we started experimenting with the generative models, which take a selected information from the KG, the corresponding dialogue act, and a context window as an input. Based on the information, the model is supposed to generate a natural language response. The models used in our experiments are not yet ready for a production-level system, but we want to continue with the experiments, and we believe that it helps the system to handle the ``long-tail'' cases in combination with template-based responses.

\medskip

\small
\printbibliography

\appendix
\section{Dialogue Acts}\label{ap:da}

\scalebox{0.8}{%
\begin{tabular}{ | l | l | l | l | l | l | }
\hline
	\textbf{T1} & \textbf{T2} & \textbf{T3} & \textbf{Description} & \textbf{Corresponding SWBD tags} &  \textbf{Example} \\ \hline \hline
	da &  &  &  &  &   \\ \hline
	 & Que &  & request for information, ``question'' &  &  \\ \hline
	 &  & Yesno & request for either confirmation or denial & qy, qy\^d, \^d & Do you like music? \\ \hline
	 &  & Wh & request for specific information & qw, qw\^d &  \\ \hline
	 &  & WhOb & question on gramm. object & & Where were you born? \\ \hline
	 &  & WhSub & question on gramm. subject & & Who starred in Matrix? \\ \hline
	 &  & Choice & request for choosing between a set options & qrr &  \\ \hline
	 &  & Howabout & contextual request for information & qo &   \\ \hline
	 & Ans &  & Answer/reaction &  &  \\ \hline
	 &  & Affirm & affirmative reaction & ny, na, aa, aap\_am & Yes, I do. \\ \hline
	 &  & Deny & negative reaction & nn, ng, ar, arp\_nd &No, he wasn't. \\ \hline
	 &  & Agree & agreeing reaction & aa, aap\_am &  \\ \hline
	 &  & Refuse & refusal, disagreeing reaction & ar, arp\_nd & I'd rather not. \\ \hline
	 &  & Suspend & suspended reaction & \^h & I'll have to think about it \\ \hline
	 &  & Tosummons & indication of presence / conversation can go on &  &\\ \hline
	 &  & Clash & ``oh really?'' ``wow!'' &  & Oh really?  \\ \hline
	 & Inf & & Providing information &  &   \\ \hline
	 &  & Obj & objective information & sd, no & John has three siblings. \\ \hline
	 &  & Subj & subjective information & sv, \^q, no & I really like sports.\\ \hline
	 &  & Repeat & repeating previous information & b\^m &\\ \hline
	 &  & Clarif & clarification, further explanation, correction & sd, no, sv, \^q, b\^m & \\ \hline
	 & Act & & Non-verbal action, stopping, volume up etc. &  &  \\ \hline
	 & Req & & Request for verbal/non-verbal action &  & \\ \hline
	 &  & Repeat & request for repeat & br & \\ \hline
	 &  & Clarif & request for clarification &  & \\ \hline
	 &  & Summons & summons &  &\\ \hline
	 &  & Action & command, request for a non-verbal action & ad & \\ \hline
	 &  & Verif & req for verification (via summary/reformulation) & bf & \\ \hline
	 &  & Sugg & suggestion, proposal & ad & \\ \hline
	 & Cont & & Continuators &  &  \\ \hline
	 &  & Rhet & rhetorical question & qh & \\ \hline
	 &  & Hm & ``uh-huh'' & b, ba & Uh-huh? \\ \hline
	 &  & Que & ``like really?!!'' & bh &  \\ \hline
	 &  & Ackn & response aknowledgement & bk & Oh, okay.\\ \hline
	 &  & Collab & collaborative completion & \^2 & \\ \hline
	 & Form & & Formalities &  &  \\ \hline
	 &  & Hello & greeting & fp &  \\ \hline
	 &  & Bye & farewell & fc & \\ \hline
	 &  & Open & conventional opening & fp & \\ \hline
	 &  & Close & conventional closing & fc & \\ \hline
	 &  & Thx & thanking & ft & Thanks a lot! \\ \hline
	 &  & Sorry & apology & fa & I am sorry\ldots \\ \hline
	 &  & Nw & downplaying response & bd & Nah, it's no problem. \\ \hline
	 & Inv & & Invalid act &  & \\ \hline
	 &  & Inv & abandoned, uniterpretable, nonverbal & \%, x &  \\ \hline
	 &  & Other & other & fo\_o\_fw\_by\_bc, t1, t3 & \\ \hline
\end{tabular}
}

\end{document}